%% file: paper.tex
\documentclass[11pt]{article}
\usepackage{stylefiles/coling20/coling2020}
\usepackage{times}
\usepackage{url}
\usepackage{latexsym}

\input{custom_preamble.tex}

\title{Learning Emotion from 100 Observations: Unexpected Robustness of  Deep Learning under Strong Data Limitations}

 \author{Sven Buechel$^{*}$ \\
   Friedrich-Schiller-Universit\"at Jena \\
   {\tt sven.buechel@uni-jena.de} \\
   \And
   Jo\~{a}o Sedoc\\
   New York University \\
   {\tt jsedoc@stern.nyu.edu} \\
   \AND
   H. Andrew Schwartz\\
   Stony Brook University\\
   {\tt has@cs.stonybrook.edu}\\
   \And
   Lyle Ungar\\
   University of Pennsylvania\\
   {\tt ungar@cis.upenn.edu}\\
 }

\date{}

\begin{document}
\maketitle
\begin{abstract}
One of the major downsides of Deep Learning is its supposed need for vast amounts of training data. As such, these techniques appear ill-suited for NLP areas where annotated data is limited, such as less-resourced languages or emotion analysis, with its many nuanced and hard-to-acquire annotation formats. We conduct a questionnaire study indicating that indeed the vast majority of researchers in emotion analysis deems neural models inferior to traditional machine learning when training data is limited. In stark contrast to those survey results, we provide empirical evidence for English, Polish, and Portuguese that commonly used neural architectures can be trained on surprisingly few observations, outperforming $n$-gram based ridge regression on only 100 data points. Our analysis suggests that high-quality, pre-trained word embeddings are a main factor for achieving those results.
\end{abstract}

\section{Introduction}
\label{sec:intro}

\blfootnote{
    %
    %
    %
    %
    %
    %
    \hspace{-0.65cm}  
    This work is licensed under a Creative Commons 
    Attribution 4.0 International License.
    License details: 
    \url{http://creativecommons.org/licenses/by/4.0/}.
}

\blfootnote{* Work partially conducted at the University of Pennsylvania.}

Deep Learning (DL) has radically changed the rules of the game in NLP by boosting performance figures in almost all application areas. Yet in contrast to more conventional techniques, such as $n$-gram based linear models, \textit{neural} methodologies seem to rely on vast amounts of training data, as is obvious in areas such as machine translation \cite{Vaswani17} or representation learning for individual words \cite{Mikolov13nips,Pennington14} or contextualized word sequences \cite{Devlin18bert,Yang19xlnet,Joshi19spanbert}.

With this profile, DL seems ill-suited for many prediction tasks in sentiment and subjectivity analysis \cite{Balahur14}.
For the widely studied problem of polarity prediction  (distinguishing only between \textit{positive} and \textit{negative} emotion), training data is relatively abundant especially for the social media domain \cite{Rosenthal17}. However, in recent years, there has been a growing interest in more nuanced and informative annotation formats for affective states \cite{Bostan18coling,DeBruyne19arxiv}. Such annotation schemes often follow distinct psychological theories such as the dimensional approach to emotion representation \cite{Bradley94} or basic emotions \cite{Ekman92}. Yet, annotating for more complex representations of affective states seems to be significantly harder in terms of both time consumption and inter-annotator agreement (IAA) \cite{Strapparava07}. Adding even more complexity, computational work following this trend often uses numerical scores as target variables making, emotion analysis a regression, rather than a classification problem \cite{Buechel16ecai,Mohammad18semeval}. 
What makes this situation even worse is that, first, we currently have a situation where there is no community-wide consensus on how emotion should be represented. That is, different ways of annotating emotion (see, e.g., Table \ref{tab:examples}) compete with each other, leading to decreased inter-operability of language resources and provoking additional data sparsity \cite{buechel18lrec}. And, second, especially large-scale annotated corpora are almost exclusively available for English, leaving most of the world's languages with little or no gold data at all. 

For the social media domain, this lack of gold data can be partly countered by (pre-)training with distant supervision using signals such as emojis or hashtags as a surrogate for manual annotation \cite{Mohammad15,Felbo17emnlp,Abdul17}. Yet, this procedure is less viable for target domains other than social media, as well as for predicting other subjective phenomena such as empathy, uncertainty, or personality  \cite{Khanpour17,Rubin07,Liu17personality}. 
Besides pre-training the entirety of the model with distant supervision, an alternative strategy is pre-training word representations, only. 
This approach is feasible for a wide range of languages, including otherwise less-resourced ones, since raw text is much more readily available than gold data, e.g., through Wikipedia \cite{Grave18}. Very recently, \textit{contextualized} word representations generated by pre-trained language models have established themselves as a powerful alternative \cite{Peters2018naacl,Devlin18bert}.

In summary, deep learning supposedly depends on vast amounts of annotated data---and this seems particularly troublesome for the field of emotion analysis because such phenomena are intrinsically hard to annotate. However, we suspect that this \textit{gold} data dependency may, for emotion analysis at least, be less severe than anticipated because large, pre-trained embedding models already seem to encode word-level emotion quite well \cite{Du16,Li17,Buechel18naacl}, possibly allowing to fit sentence-level DL architectures on rather small datasets. If that was true, it would be the \textit{reputation} of DL rather than its actual characteristics which prevent its wider application for emotion analysis in low-resource environments.

\textbf{Contribution. } We start by quantifying the expectations of the research community regarding the data requirements of DL. To this end, we first conduct a questionnaire study among NLP researchers in the field of emotion analysis finding that the median respondent expects DL to be viable only from 10,000 training examples onward. Next, we perform a series of experiments on English, Polish, and Portuguese emotion corpora.  In contrast to the survey results, we show that commonly used architectures can be fitted on as little as 100 data points and still outperform supposedly more robust $n$-gram based approaches. We believe these findings potentially open up DL to many low-resource areas and, by extension, wider cross-lingual or cross-domain applications. 




\section{Survey}

\begin{figure}
  \centering
  \small
  \begin{mdframed}
  \textbf{Data Requirements for Deep Learning in Emotion Analysis}

  We're interested in your beliefs about what training size is necessary for deep learning techniques. Your response will be used for academic research and stored anonymously. Thank you very much!

  Consider the task of fine-grained emotion analysis, using what you believe to be the best deep learning architecture (e.g., RNN, CNN, Self-Attention) with input from pre-trained word embeddings (e.g., word2vec, GloVe; NOT contextualized embeddings like ELMO or BERT):
  \begin{enumerate}[noitemsep]
    \item How many training examples do you believe are necessary for deep learning to provide results in line with traditional discriminative learning (e.g. SVM, penalized linear regression, random forests)?
    \item How many observations do you believe are necessary for deep learning approaches to provide a clear benefit over traditional discriminative learning?
  \end{enumerate}
  Thank you for completing the survey! Do you have any additional comments regarding this questionnaire?
  \end{mdframed}
  \caption{\label{fig:surveytext} Survey on expected data requirements of deep learning.}
\end{figure}

\input{figs/survey.tex}

 We conducted a short questionnaire study asking the research community about their beliefs regarding data requirements of deep learning in the context of emotion analysis. We included two questions, one asking for the number of training examples ``necessary for deep learning to provide results\textit{ in line} with traditional discriminative learning'' (question 1), the other asking for the number of examples necessary for ``deep learning approaches to provide a \textit{clear benefit} over traditional discriminative learning'' (question 2). The full text of the questionnaire is given in Figure \ref{fig:surveytext}. Participants were instructed to focus on \textit{non}-contextualized word representations (in line with our latter experiments; the use of contextualized word embeddings is left for future work) being used as input to the, from their view, most suitable model architecture.

To recruit participants, we queried the ACL Anthology for papers from between 2016 and 2018 using the keyword ``emotion'', we collected all email addresses in the author field of all retrieved PDFs (166 papers in total).
Invitations to participate in the survey were sent to the resulting 391 email addresses on February 28, 2019. We received 26 responses within four weeks (6\% response rate).  One response (stating in the optional comment field that no numeric answer could be given) was excluded. Figure \ref{fig:survey} shows the distribution of the 25 remaining responses on a logarithmic scale.

As can be seen, the responses to both questions clearly support the intuition that deep learning is thought of as being dependent on large amounts of training data by the scientific community. In both cases, the median response was 10,000. Perhaps surprisingly, a total of 5 participants stated that fewer than 100 instances are necessary for deep learning to show clear improvements over more traditional methods (compared to 20 who believed the opposite), whereas only 2 believed that less than 100 instances are enough to show results ``in line with'' traditional methods. Inspecting the individual responses, we found that this discrepancy stems from a minority of participants (4 of out 25) who indicated that traditional learning performs worse than DL on small datasets but may catch up as dataset sizes grow.

Another interesting, most likely related outcome is that the responses to question 2 show a bi-modal distribution: While a minority of 5 participants believed that DL approaches are superior below 100 observations, the vast majority of participants (20) states that 1,000 or more instances are necessary for that. Yet, no one responded with a number between 100 and 1,000.

While we do not validate the claim of this minority, the remainder of the paper provides strong evidence that the majority of the participants largely overestimated the data requirements of deep learning.


\section{Data}
\label{sec:data}

\input{tabs/data.tex}
\input{tabs/examples_extended.tex}

For the following study, we selected four small ($<3000$ instances) and typologically diverse datasets described below.
Pre-trained, publicly available word2vec \cite{Mikolov13nips} and FastText vectors \cite{Bojanowski17} of matching language and target domain were used as model input. Table \ref{tab:data} summarizes the employed data. Illustrative examples of the particular styles and annotation formats of those corpora are provided in Table \ref{tab:examples}.

{\bf SE07}: The test set of SemEval 2007 Task 14 \cite{Strapparava07} comprises 1000 English news headlines that are annotated according to six Basic Emotions, joy, anger, sadness, fear, disgust, and surprise on a $[0, 100]$-scale (BE6 annotation format). The news headlines are quite short and rather objective, being written by professional journalists. However,  they may still elicit strong emotional reactions in readers as illustrated in Table \ref{tab:examples}. For this corpus, we used the word2vec embeddings trained on Google News.\footnote{
    \url{code.google.com/archive/p/word2vec/}
}

{\bf WASSA}: The English Twitter dataset of the WASSA 2017 shared task \cite{Mohammad17wassa} contains four subsets, one for each of the first four basic emotions, annotated on a $[0, 1]$ scale (BE4 format). Their sizes vary between 1533 and 2252 samples (union of the respective train, dev, and test set). Being a Twitter corpus, these data display features typical for the social media domain, e.g., extensive use of colloquialism, emojis, and explicit language, as well as platform-specific phenomena such as hashtags (`\#') and user mentions (`@'). Note that different from other corpora, here each individual instance is annotated according to only one emotion variable, whereas in SE07, ANPST, and MAS every instance is annotated for all variables covered by the respective dataset. Here, we used Twitter word2vec embeddings by \newcite{Godin2015}.

{\bf ANPST}: The Affective Norms for Polish Short Texts \cite{Imbir17} is a  dataset designed as stimulus in psychological experiments. It is annotated according to valence, arousal, and dominance on a $[1, 9]$-scale (VAD). ANPST comprises sentences of various genres (such as proverbs, jokes, literature quotes, or newswire material) from a wide range of sources \cite{Imbir17}.  The resulting selection of raw data seems often quite complex and ambiguous in terms of the elicited emotion (see examples in Table \ref{tab:examples}).
We used the FastText embeddings by \newcite{Grave18} trained on the Polish Wikipedia.

{\bf MAS}: Like ANPST, the Minho Affective Sentences \cite{Pinheiro17} is a dataset designed by psychologists, also being annotated according to valence, arousal, and dominance on a $[1, 9]$-scale (VAD). Yet, additionally, MAS is also annotated according to the first five Basic Emotions (omitting `surprise') on a $[1, 5]$-scale (BE5). 
It consists of very short situation descriptions in the third person in European Portuguese \cite{Pinheiro17}. Those sentences were purposefully constructed by psychologists to be simple in language and familiar in content for a large proportion of the population. This was done to make the dataset more widely applicable as experimental stimulus, yet also resulted in a slightly artificial style. MAS is the smallest dataset considered, having only 192 instances. We used the FastText embeddings by \newcite{Grave18}, trained on the Portuguese Wikipedia.

\section{Methods}
\label{sec:methods}

We provide two distinct linear baseline models which both rely on Ridge regression, an $\ell^2$-regularized version of linear regression. The first one, Ridge\textsub{ngram}, is based on $n$-gram features where we use $n \in \{1,2,3\}$. The second one, Ridge\textsub{BV} uses \textit{bag-of-vectors} features, i.e., the pointwise mean of the embeddings of the words in a text.
Regarding the deep learning approaches, we compare Feed-Forward Networks (FFN), Gated Recurrent Unit Networks (GRU), Long Short-Term Memory Networks (LSTM), Convolutional Neural Networks (CNN), as well as a combination of the latter two (CNN-LSTM) \cite{Cho14gru,Hochreiter97,Kalchbrenner14}.

Since holding out a dev set from the already limited training data does not seem feasible for some of the datasets (see Table \ref{tab:data}), we decided to instead use constant hyperparameter settings across all corpora. We also keep most hyperparameters constant between models. Hence, hyperparameter choices followed well-established recommendations described in the next paragraph.
%
%
%

%

\input{tabs/models.tex}

The input to the DL models is based on pre-trained word vectors. ReLu activation was used everywhere except in recurrent layers. Dropout is used for regularization with a probability of .2 for embedding layers and .5 for dense layers following the recommendations by \newcite{Srivastava14}. We use .5 dropout also on other types of layers where it would conventionally be considered too high (e.g. on max pooling layers). Our models are trained for 200 epochs using the Adam optimizer \cite{Kingma15} with a fixed learning rate of $.001$ and a batch size of 32.
Word embeddings were not updated during training.
Since, in compliance with our gold data, we treat emotion analysis as a regression problem \cite{Buechel16ecai},
the output layers of our models consist of an affine transformation, i.e., a dense layer without non-linearity.
To reduce the risk of overfitting on such small data sets, we used relatively simple models both in terms of the number of layers and units in them (mostly 2 and 128, respectively). An overview of our models and  details about their \textit{individual} hyperparameter settings are provided in Table \ref{tab:models}. \url{Keras.io}  and \url{scikit-learn.org} \cite{Pedegrosa11} were used for the implementation.

\section{Experimental Results}
\label{sec:results}

\input{tabs/cv_results.tex}

\input{figs/training_size.tex}
\input{tabs/embedding_strategies.tex}
\input{tabs/semeval.tex}
\input{tabs/wassa.tex}

\subsection{Repeated Cross-Validation}
Given our small datasets, conventional 10-fold cross-validation (CV) would lead to very small test splits (only 19 instances in the case of MAS) thus causing high variance between the individual splits and,  ultimately, even regarding the average of all 10 runs. Therefore, we {\it repeat 10-fold CV ten times} ($10{\times}10$-CV) with different data splits, then averaging the results \cite{Dietterich98}. Performance is measured as Pearson correlation $r$ between predicted and human gold ratings. To further increase reliability,  identical data splits were used for each of the approaches under comparison. Results are given in Table \ref{tab:cv_results}.

All DL approaches (FFN, CNN, GRU, LSTM, CNN-LSTM) yield a satisfying performance of $r>.6$ on average over all corpora, despite the small data sizes. Each one of them clearly outperforms Ridge\textsub{ngram}, representing more conventional learning techniques, on every single dataset. This stands in sharp contrast to our survey results where the median respondent indicated that DL would need at least 10,000 instances to provide a clear benefit over conventional techniques---the datasets we employed are between 4 and 50 times smaller.

Overall, the GRU performs best. However,  differences between the DL models are quite small on average. (We emphasize that our primary concern is to compare deep vs. conventional learning techniques under data limitations whereas comparisons within the group of DL architectures are secondary.)
%
Perhaps surprisingly, Ridge\textsub{BV}, which takes a middle ground between DL and conventional approaches,  also performs very competitively. Given its low computational cost, our results indicate that this model may constitute an excellent baseline. 

Observe that Ridge\textsub{BV} and FFN rely solely on lexical information---their inputs are computed by mere averaging of word embeddings whereas CNN, LSTM, GRU, and CNN-LSTM learn their own  composition functions from gold data. Still, the former two both display satisfying performance. This suggests that the quality of the pre-trained embeddings may be a key factor for their strong results.

\subsection{Embedding Training Strategies} To further examine this conjecture, we repeated the above experiment two more times, altering the training strategy of the embeddings (only applicable to DL models).
%
%
%
Instead of using pre-trained vectors without updating them (\textbf{Frozen}), we looked at embeddings which were either randomly initialized and updated (\textbf{Learned}) or pre-trained and updated (\textbf{Tuned}).
As can be seen from Table \ref{tab:strategies}, both strategies involving pre-trained vectors (frozen and tuned) outperform learned word embeddings by a large margin (about $30\%$-points on average).
Frozen embeddings yield the highest performance, even outperforming fine-tuned vectors ($5\%$-point margin on average), a possible reason being that the large increase in the number of parameters leads to overfitting.

\subsection{Training Size vs. Model Performance}
We will now continue to explore the unexpected behavior of DL architectures by further limiting the available training data.
For each number $N{\in}\{1, 10, 20, ..., 100, 200,..., 900\}$, we randomly sampled $N$ instances from the SE07 corpus for training and tested on the held-out data. This procedure was repeated 100 times for each of the training data sizes before averaging the results. Each of the models was evaluated with identical data splits. The outcome of this experiment is depicted in Figure \ref{fig:training_size}.
As can be seen, recurrent models suffer only a moderate loss of performance down to a third of the original training data (about 300 observations).
The CNN, FFN, and Ridge\textsub{BV}  models remain stable even longer---their performance only begins to decline rapidly at about 100 instances. 
In contrast, Ridge\textsub{ngram} declines more steadily yet its overall performance is much lower as well.
Most notably, all DL models but the LSTM \textit{always} performed better than the conventional Ridge\textsub{ngram} baseline no matter how little training data was used.

\subsection{Comparison against Previous Work}

The above experiments have shown that our DL models perform robustly under strong data limitations, beating a conventional baseline in the vast majority of cases.  Yet, perhaps this was achieved by designing overly simple network architectures, thus trading an excessive amount of performance for robustness in low-data scenarios. To rule out this possibility, we will now move forward and compare our findings against previous work.


\paragraph{SemEval 2007 Affective Text}

First, we compare our best performing model, the GRU, against previously reported results for the SE07 corpus. Table \ref{tab:semeval} provides the performance of the winning system of the original shared task ({\sc Winner}; \newcite{Chaumartin07}), the inter-annotator agreement (IAA) as given by the organizers \cite{Strapparava07}, the performance by \newcite{Beck17}, the highest one reported for this dataset so far ({\sc Beck}), as well as the results for our GRU from the $10{\times}10$-CV set-up.

As can be seen, the GRU established a new state-of-the-art surpassing the previous one by about 4\%-points on average over all emotion categories. The difference is statistically significant (two-tailed one-sample $t$-test comparing the results of the 10 cross-validation runs against the reported performance by \newcite{Beck17}; $p<.001$).
Our GRU also outperforms IAA, as already did \textsc{Beck}. This may sound improbable at first glance. However, \newcite{Strapparava07} employ a rather weak notion of human performance which is---broadly speaking---based on the reliability of a single human rater.\footnote{
    Instead, other approaches to IAA computation for numerical values, such as split-half or inter-study reliability, constitute a more challenging comparison since they are based on the reliability of \textit{many} raters, not one \cite{Mohammad17starsem,Buechel18coling}.
}
Interestingly, the GRU shows particularly large improvements over human performance for categories where the IAA is low (anger, disgust, and surprise).

\paragraph{WASSA 2017 Shared Task Data}
Table \ref{tab:wassa} displays the official results of the four best systems (out of 21 submissions) of the WASSA 2017 shared task \cite{Mohammad17wassa} as well as the performance our GRU achieved.
%
For this experiment, we deviated from the above $10{\times}10$-CV set-up but instead used the official train-dev-test split for comparability. 
As for all experiments in this paper, hyperparameters were kept constant and were not adjusted to this dataset. Consequently, train and dev sets were combined for training. Training and testing were repeated ten times with different random seeds but otherwise identical configuration following the recommendation by \newcite{Reimers18arxiv}.  Table \ref{tab:wassa} shows our \textit{average} performance over those ten runs.

As can be seen, our GRU performs very competitively and would have been ranked fourth place, outperforming 18 out of 21 submissions. 
The difference to the next lower-performing system (UWaterloo) is statistically significant (two-tailed one-sample $t$-test comparing our ten runs against their official results; $p<.001$).


\section{Conclusion}
Annotating emotion is necessarily subjective thus making gold data in this area particularly rare. As such, applying DL may seem ill-advised since supposedly large amounts of training data are required. But is this really the case?
We started our investigation by conducting a survey among researchers in emotion analysis. 80\% of the respondents believed that DL is superior to traditional machine learning techniques only when at least 1,000 training examples are available. Half of the participants even believed that 10,000 or more examples are necessary.
Putting this popular notion to the test, we provided the first examination of neural emotion analysis under severe data constraints, featuring five distinct neural architectures and three typologically diverse languages.
In stark contrast to the survey results, we found that \textit{all} architectures could be fitted on datasets comprising as little as 200 observations, CNNs and FFNs even being robust on 100 observations.
A subsequent analysis indicated that
high-quality, pre-trained word embeddings are a key factor in achieving those results. 
%
%
%
In the future, we would like to extend this work to \textit{contextualized} word representations, e.g., by ELMo or BERT \cite{Peters2018naacl,Devlin18bert}.
%

%

 \section*{Acknowledgements}
 
We thank the anonymous reviewers for their helpful suggestions and comments. Sven Buechel thanks his doctoral advisor Udo Hahn, JULIE Lab,  for funding his research visit at the University of Pennsylvania.

\bibliographystyle{stylefiles/coling20/acl}
\bibliography{literature}

\end{document}

%% file: custom_preamble.tex
\usepackage[T1]{fontenc}
\usepackage{mathtools}
\usepackage{multirow}
\usepackage{subcaption}
\usepackage{amsmath}
\usepackage{amssymb}
\usepackage{booktabs}
\usepackage{rotating}
\usepackage{enumitem}
\usepackage{todonotes}
\usepackage[linewidth=1pt]{mdframed}

\newcommand{\textsub}[1]{$_\text{#1}$}
\newcommand{\rt}[1]{\begin{rotate}{30}#1\end{rotate}}

%% file: figs/survey.tex
\begin{figure}[t]
    \begin{minipage}{.5\textwidth}
    \centering
    \includegraphics[width=.95\textwidth]{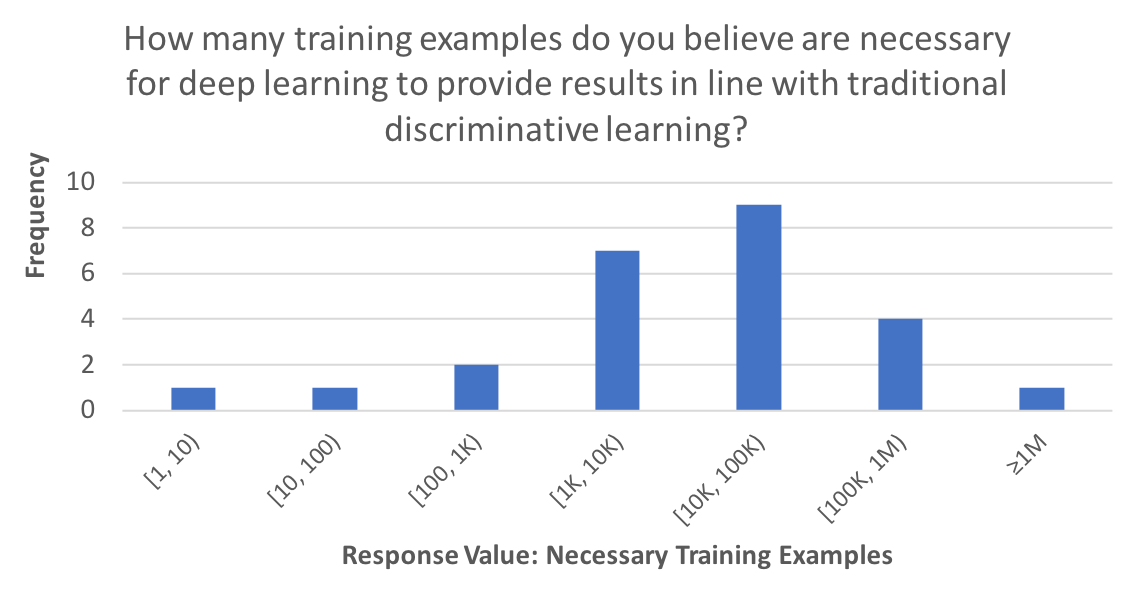}
    \end{minipage}
    \begin{minipage}{.5\textwidth}
    \centering
    \includegraphics[width=.95\textwidth]{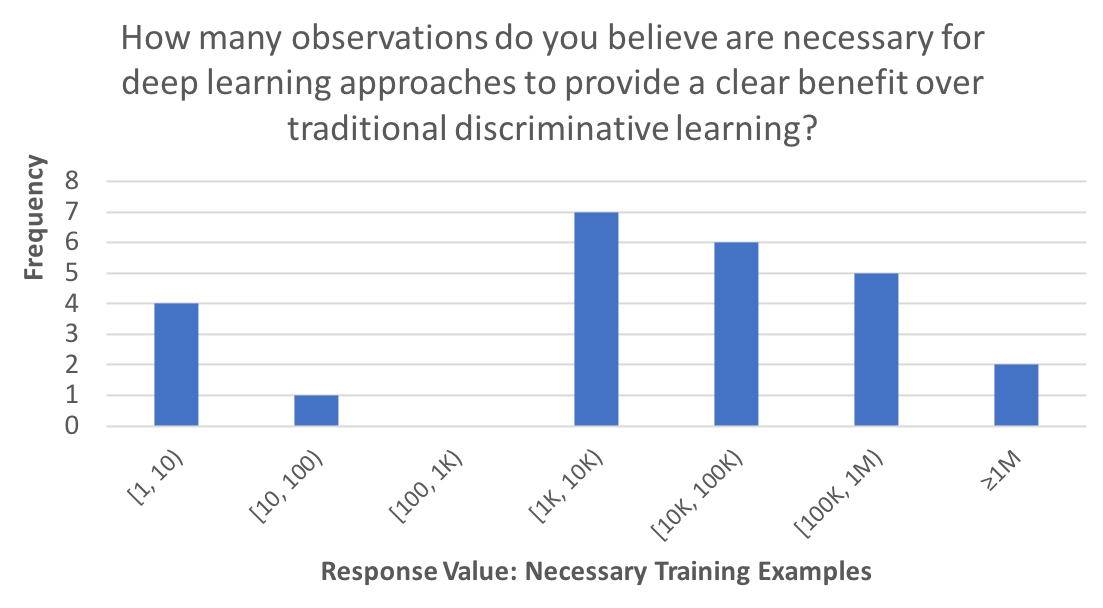}
    \end{minipage}
    \caption{\label{fig:survey}Responses to survey questions 1 (left) and 2 (right).}
\end{figure}

%% file: tabs/data.tex
\begin{table*}[t!]
    \centering
    \small
    \begin{tabular}{llrlrrlr}
    \toprule
        {\bf Corpus} & {\bf Language} & {\bf Size} & {\bf Annotation} & {\bf Emb. Alg.} & \textbf{Dims.} & {\bf Emb. Data} &  {\bf Emb. Data Size}\\
        \midrule
        SE07 & English & 1000 & BE6 $[1, 100]$ & word2vec & 300 & Google News  & 100B tokens\\
        WASSA & English & $\leq 2252$& BE4 $[0, 1]$ & word2vec & 400 & Twitter & 400M tweets \\
        ANPST & Polish & 718& VAD $[1, 9]$ & FastText & 300 & Wikipedia & 4B tokens\\
        MAS & Portuguese & 192 &  VAD $[1, 9]$ + BE5 $[1,5]$ & FastText & 300& Wikipedia  & 4B tokens \\
        \bottomrule
    \end{tabular}
    \caption{Annotated corpora and embedding models used for experiments; with language, number of instances, annotation format, embedding algorithm, embedding dimensions, and dataset (size) embeddings were trained on.}
    \label{tab:data}
\end{table*}

%% file: tabs/examples_extended.tex
\begin{table*}[t!]
    \centering
    \small
    \begin{tabular}{lp{5.5cm}rrrrrrrrr}
    \toprule
    \textbf{Corpus} & \textbf{Text} & \textbf{Val} & \textbf{Aro} & \textbf{Dom} & \textbf{Joy} & \textbf{Ang} & \textbf{Sad} & \textbf{Fea} & \textbf{Dis} & \textbf{Sur}\\
    \midrule
    \multirow{3}{*}{SE07}     & \textit{Inter Milan set Serie A win record} & - & - & - & 50 & 2 & 0 & 0 & 0 & 9 \\
    \cmidrule{2-11}
     & \textit{TBS to pay \$2M fine for ad campaign bomb scare} & - & - & - & 11 & 25 & 28 & 45 & 32 & 43 \\
    \midrule
    \multirow{3}{*}{WASSA} & \textit{@TheRevAl please tell us why 'protesting' injustice requires \#burning \#beating and \#looting terrible optics \#toussaintromain is true leader!} & - & - & - & - & .73 & - & - &- & -\\
    \cmidrule{2-11}
    &\textit{@TauDeltaPhiDK THANK YOU FOR MY OBAMA CUT OUT!!!!!! I am elated that he's back home\includegraphics[]{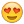}}& - & - & - & .83 & - & - & - & - & -\\
    \midrule
    \multirow{3}{*}{ANPST}     & \textit{Decyzje podj\k{e}te w przesz\l{}o\'{s}\'{c} i kszta\l{}tuj\k{a} nasz\k{a} tera\'zniejszo\'{s}\'{c}.} `Decisions made in the past shape our present.' & 4.9 & 4.1 & 5.8 & - & - & - & - & - & - \\
    \cmidrule{2-11}
    & \textit{Dop\'{o}ki walczysz i podejmujesz starania, jeste\'{s} zwyci\k{e}zc\k{a}.} `As long as you fight and keep trying, you are a winner.' & 7.3 & 5.3 & 7.4 & - & -& -& -& -& -\\
    \midrule
    \multirow{3}{*}{MAS}     &  \textit{A praia \'e espetacular.} `The beach is spectacular.' & 8.0 &	4.0 &	6.7 & 4.2 & 1.0 & 1.0 & 1.0 & 1.0 & -\\
    \cmidrule{2-11}
    & \textit{A tinta \'{e} azul.} `The ink is blue.' & 5.0 & 3.9 & 5.5 & 1.3 & 1.0 & 1.0 & 1.0 & 1.0 & - \\
    \bottomrule

    \end{tabular}
    \caption{Exemplary entries from our four datasets illustrating differences in linguistic characteristics and emotion annotation scheme. Emotion variables: valence, arousal, dominance, joy, anger, sadness, fear, disgust, and surprise. English translations for ANPST and MAS were provided by the respective dataset creators.}
    \label{tab:examples}
    \vspace{-6pt}
\end{table*}

%% file: tabs/models.tex
\begin{table}
    \centering
    \small
    \begin{tabular}{p{1.7cm}p{4.9cm}}
    \toprule
    {\bf Model} &  {\bf Description}\\
    \midrule
     Ridge\textsub{ngram} & $n$-gram features with $n\in\{1,2,3\}$; feature normalization; automatically chosen regularization coefficient from $\{10^{-4}, 10^{-3}, .., 10^4\}$\\
     \midrule
     Ridge\textsub{BV}& \textit{bag of vectors}-features; regularization coefficient chosen as in `Ridge\textsub{ngram}'\\
    \midrule
     FFN    & \textit{bag of vectors}-features; two dense layers (256 and 128 units)\\
     \midrule
     CNN    & one conv. layer (filter size 3, 128 channels), max-pooling layer with .5 dropout; dense layer (128 units) \\
     \midrule
     GRU    & recurrent layer (128 units, uni-directional); last timestep receives .5 vertical dropout and is fed into a dense layer (128 units)\\
     \midrule
     LSTM   & identical to `GRU'\\
     \midrule
     CNN-LSTM & conv. layer as in `CNN'; max-pooling layer (pool size 2, stride size 1) with .5 dropout; LSTM identical to `GRU'\\
    \bottomrule
    \end{tabular}
    \caption{Model-specific design choices.}
    \label{tab:models}
\end{table}

%% file: tabs/cv_results.tex
\begin{table}[t]
    \centering
    \small
\vspace*{30pt}
\begin{tabular}{lccccc}                  
{} &  \rt{\textbf{SE07}} &  \rt{\textbf{WASSA}} &  \rt{\textbf{ANPST}} &   \rt{\textbf{MAS}} &  \rt{\textbf{Mean}} \\               
\midrule
Ridge\textsub{ngram} &     .53 &   .67 &   .32 &  .16 &  .42 \\                 
Ridge\textsub{BV}    &     .62 &   .64 &   .52 &  .62 &  .60 \\ 
CNN-LSTM &     .66 &   .69 &   .50 &  .63 &  .62 \\
CNN     &     \textbf{.67} &   .70 &   .47 &  .61 &  .62 \\
FFN     &     .67 &   .69 &   .50 &  .65 &  .63 \\
LSTM    &     .65 &   .73 &   .52 &  .65 &  .64 \\                   
GRU     &     .67 &   \textbf{.73} &   \textbf{.54} &  \textbf{.66} &  \textbf{.65} \\          
\midrule
\end{tabular}
\caption{Comparative results of the $10{\times}10$-cross-validation in Pearson's $r$; averaged over all variables of the respective annotation format.}
\label{tab:cv_results}
\end{table}

%% file: figs/training_size.tex
\begin{figure}[t]
    \centering
    \includegraphics[width=.43\textwidth]{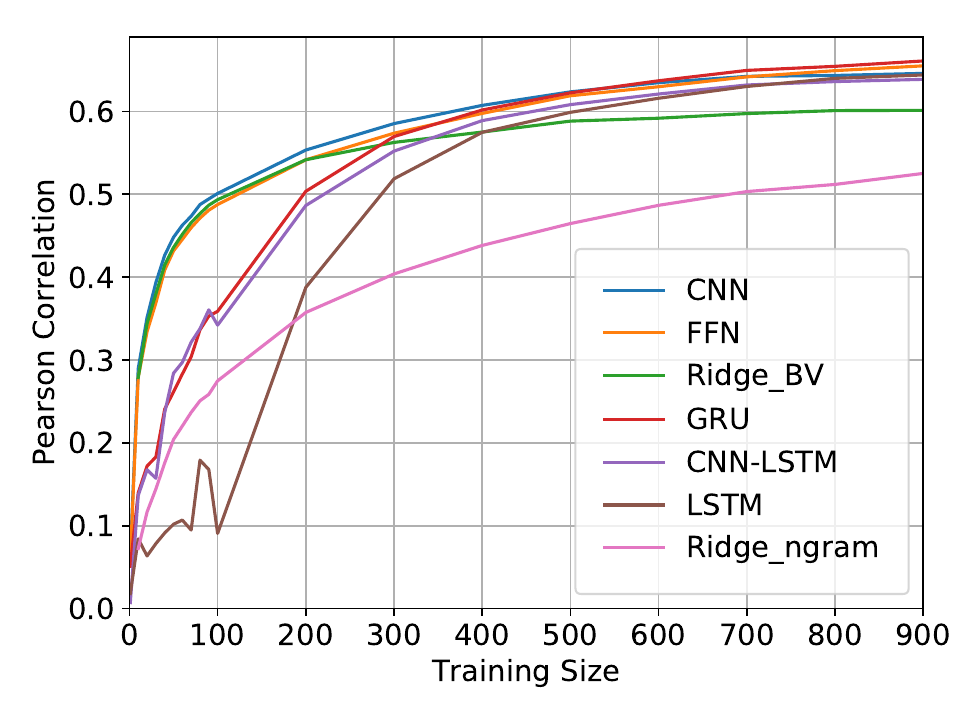}
    \caption{Comparison of model performance vs. training size on the SE07 dataset in Pearson's \textit{r}.}
    \label{fig:training_size}
\end{figure}

%% file: tabs/embedding_strategies.tex
\begin{table}[t]
\small
\center
\vspace*{.8cm}
\begin{tabular}{lccccccc}
{} &   \textbf{\rt{FFN}} &   \textbf{\rt{CNN}} &   \textbf{\rt{GRU}}&  \textbf{\rt{LSTM}} &  \textbf{\rt{CNN-LSTM}} &  \textbf{\rt{Mean}} \\
\midrule
\textbf{Learned} &  .24 &  .23 &  .38 &  .26 &     .30 &  .28 \\
\textbf{Tuned}   &  .59 &  .55 &  .59 &  .59 &     .57 &  .58 \\
\textbf{Frozen}  &  \textbf{.63} &  \textbf{.62} &  \textbf{.65} &  \textbf{.64} &     \textbf{.62} &  \textbf{.63} \\
\midrule
\end{tabular}
\vspace*{-6pt}
\caption{\label{tab:strategies}
Comparison of embedding training strategies (average Pearson's $r$ over all datasets).
}
\end{table}

%% file: tabs/semeval.tex
\begin{table}[b!]
    \centering
    \small
\vspace{10pt}
\begin{tabular}{lccccccc}
{} &  \textbf{\rt{Joy}} &  \textbf{\rt{Anger}} &  \textbf{\rt{Sadness}} &  \textbf{\rt{Fear}} &  \textbf{\rt{Disgust}} &  \textbf{\rt{Surprise}} &   \textbf{\rt{Mean}}   \\
\midrule
{\sc Winner}                         &  .23 &    .32 &      .41 &   .45 &      .13 &       .17 & .28 \\
IAA                          &  .60 &    .50 &      .68 &   .64 &      .45 &       .36 & .54  \\
{\sc Beck}              &  .59 &    .65 &      .70 &   .74 &      .54 &       .47 & .62 \\
GRU                      &  {\bf.60} &    {\bf.70} &      {\bf.75} &  {\bf .77} &      {\bf.61} &  {\bf.53} & {\bf.66}  \\
\midrule
\end{tabular}
\caption{Comparison of previously reported results, human performance (IAA), and our proposed GRU model on the SE07 dataset in Pearson's $r$.}
\label{tab:semeval}
\end{table}

%% file: tabs/wassa.tex
\begin{table}[b!]
    \centering
    \small
\vspace{30pt}
\begin{tabular}{p{.5cm}lccccc}
\textbf{\rt{Official Rank}} & \textbf{\rt{Team/System}} & \textbf{ \rt{Joy}} &  \textbf{\rt{Anger}} &  \textbf{\rt{Sadness}} &  \textbf{\rt{Fear}} & \textbf{\rt{Mean}}   \\
\midrule
1 & Prayas & .762 & .765 & .732 & .732 & .747\\
2 & IMS & .726 & .767 & .690 & .705 & .722 \\
3 & SeeNet & .698 & .745 & .715 & .676 & .708 \\
-- & \textbf{Our Work} & .658 & .668 & .724 & .717 & .692\\
4 & UWaterloo & .699 & .703 & .693 & .643 & .685\\
\midrule
\end{tabular}
\caption{Comparison against official WASSA 2017 shared task results (in Pearson's $r$).}
\label{tab:wassa}
\end{table}

%% file: paper.bbl
\begin{thebibliography}{}

\bibitem[\protect\citename{Abdul-Mageed and Ungar}2017]{Abdul17}
Muhammad Abdul-Mageed and Lyle Ungar.
\newblock 2017.
\newblock {EmoNet}: {Fine}-grained emotion detection with gated recurrent
  neural networks.
\newblock In {\em Proceedings of the 55th {Annual} {Meeting} of the
  {Association} for {Computational} {Linguistics} ({Volume} 1: {Long}
  {Papers})}, pages 718--728.

\bibitem[\protect\citename{Balahur \bgroup et al.\egroup }2014]{Balahur14}
Alexandra Balahur, Rada Mihalcea, and Andr\'{e}s Montoyo.
\newblock 2014.
\newblock Computational approaches to subjectivity and sentiment analysis:
  {Present} and envisaged methods and applications.
\newblock {\em Computer Speech \& Language}, 28(1):1--6.

\bibitem[\protect\citename{Beck}2017]{Beck17}
Daniel Beck.
\newblock 2017.
\newblock Modelling representation noise in emotion analysis using gaussian
  processes.
\newblock In {\em Proceedings of the 8th International Joint Conference on
  Natural Language Processing (Volume 2: Short Papers)}, pages 140--145.

\bibitem[\protect\citename{Bojanowski \bgroup et al.\egroup
  }2017]{Bojanowski17}
Piotr Bojanowski, Edouard Grave, Armand Joulin, and Tom\'{a}\v{s} Mikolov.
\newblock 2017.
\newblock Enriching word vectors with subword information.
\newblock {\em Transactions of the Association for Computational Linguistics},
  5(1):135--146.

\bibitem[\protect\citename{Bostan and Klinger}2018]{Bostan18coling}
Laura-Ana-Maria Bostan and Roman Klinger.
\newblock 2018.
\newblock An analysis of annotated corpora for emotion classification in text.
\newblock In {\em Proceedings of the 27th International Conference on
  Computational Linguistics}, pages 2104--2119.

\bibitem[\protect\citename{Bradley and Lang}1994]{Bradley94}
Margaret~M. Bradley and Peter~J. Lang.
\newblock 1994.
\newblock Measuring emotion: The {Self-Assessment Manikin} and the semantic
  differential.
\newblock {\em Journal of Behavior Therapy and Experimental Psychiatry},
  25(1):49--59.

\bibitem[\protect\citename{Buechel and Hahn}2016]{Buechel16ecai}
Sven Buechel and Udo Hahn.
\newblock 2016.
\newblock Emotion analysis as a regression problem: Dimensional models and
  their implications on emotion representation and metrical evaluation.
\newblock In {\em Proceedings of the 22nd European Conference on Artificial
  Intelligence}, pages 1114--1122.

\bibitem[\protect\citename{Buechel and Hahn}2018a]{Buechel18coling}
Sven Buechel and Udo Hahn.
\newblock 2018a.
\newblock Emotion representation mapping for automatic lexicon construction
  (mostly) performs on human level.
\newblock In {\em Proceedings of the 27th International Conference on
  Computational Linguistics}, pages 2892--2904.

\bibitem[\protect\citename{Buechel and Hahn}2018b]{buechel18lrec}
Sven Buechel and Udo Hahn.
\newblock 2018b.
\newblock Representation mapping: {A} novel approach to generate high-quality
  multi-lingual emotion lexicons.
\newblock In {\em {Proceedings} of the 11th {International} {Conference} on
  {Language} {Resources} and {Evaluation}}, pages 184--191.

\bibitem[\protect\citename{Buechel and Hahn}2018c]{Buechel18naacl}
Sven Buechel and Udo Hahn.
\newblock 2018c.
\newblock Word emotion induction for multiple languages as a deep multi-task
  learning problem.
\newblock In {\em Proceedings of the 2018 Conference of the North American
  Chapter of the Association for Computational Linguistics: Human Language
  Technologies, Volume 1 (Long Papers)}, pages 1907--1918.

\bibitem[\protect\citename{Chaumartin}2007]{Chaumartin07}
Fran{\c{c}}ois-R{\'e}gis Chaumartin.
\newblock 2007.
\newblock {UPAR7}: A knowledge-based system for headline sentiment tagging.
\newblock In {\em Proceedings of the 4th International Workshop on Semantic
  Evaluations}, pages 422--425.

\bibitem[\protect\citename{Cho \bgroup et al.\egroup }2014]{Cho14gru}
Kyunghyun Cho, Bart van Merrienboer, Dzmitry Bahdanau, and Yoshua Bengio.
\newblock 2014.
\newblock On the properties of neural machine translation: Encoder--decoder
  approaches.
\newblock In {\em Proceedings of SSST-8, 8th Workshop on Syntax, Semantics and
  Structure in Statistical Translation}, pages 103--111.

\bibitem[\protect\citename{De~Bruyne \bgroup et al.\egroup
  }2019]{DeBruyne19arxiv}
Luna De~Bruyne, Pepa Atanasova, and Isabelle Augenstein.
\newblock 2019.
\newblock Joint emotion label space modelling for affect lexica.
\newblock {\em arXiv:1911.08782 [cs.CL]}.

\bibitem[\protect\citename{Devlin \bgroup et al.\egroup }2019]{Devlin18bert}
Jacob Devlin, Ming-Wei Chang, Kenton Lee, and Kristina Toutanova.
\newblock 2019.
\newblock {BERT}: Pre-training of deep bidirectional transformers for language
  understanding.
\newblock In {\em Proceedings of the 2019 Conference of the North {A}merican
  Chapter of the Association for Computational Linguistics: Human Language
  Technologies, Volume 1 (Long and Short Papers)}, pages 4171--4186.

\bibitem[\protect\citename{Dietterich}1998]{Dietterich98}
Thomas~G Dietterich.
\newblock 1998.
\newblock Approximate statistical tests for comparing supervised classification
  learning algorithms.
\newblock {\em Neural Computation}, 10(7):1895--1923.

\bibitem[\protect\citename{Du and Zhang}2016]{Du16}
Steven Du and Xi~Zhang.
\newblock 2016.
\newblock Aicyber's system for {IALP 2016 Shared Task}: Character-enhanced word
  vectors and boosted neural networks.
\newblock In {\em Proceedings of the 2016 International Conference on Asian
  Language Processing}, pages 161--163.

\bibitem[\protect\citename{Ekman}1992]{Ekman92}
Paul Ekman.
\newblock 1992.
\newblock An argument for basic emotions.
\newblock {\em Cognition \& Emotion}, 6(3-4):169--200.

\bibitem[\protect\citename{Felbo \bgroup et al.\egroup }2017]{Felbo17emnlp}
Bjarke Felbo, Alan Mislove, Anders S{\o}gaard, Iyad Rahwan, and Sune Lehmann.
\newblock 2017.
\newblock Using millions of emoji occurrences to learn any-domain
  representations for detecting sentiment, emotion and sarcasm.
\newblock In {\em Proceedings of the 2017 Conference on Empirical Methods in
  Natural Language Processing}, pages 1615--1625.

\bibitem[\protect\citename{Godin \bgroup et al.\egroup }2015]{Godin2015}
Fr{\'e}deric Godin, Baptist Vandersmissen, Wesley De~Neve, and Rik Van~de
  Walle.
\newblock 2015.
\newblock {Multimedia Lab @ ACL WNUT NER Shared Task}: {Named} entity
  recognition for {Twitter} microposts using distributed word representations.
\newblock In {\em Proceedings of the Workshop on Noisy User-generated Text},
  pages 146--153.

\bibitem[\protect\citename{Grave \bgroup et al.\egroup }2018]{Grave18}
Edouard Grave, Piotr Bojanowski, Prakhar Gupta, Armand Joulin, and Tomas
  Mikolov.
\newblock 2018.
\newblock Learning word vectors for 157 languages.
\newblock In {\em Proceedings of the 11th International Conference on Language
  Resources and Evaluation}, pages 3483--3487.

\bibitem[\protect\citename{Hochreiter and Schmidhuber}1997]{Hochreiter97}
Sepp Hochreiter and J{\"u}rgen Schmidhuber.
\newblock 1997.
\newblock Long short-term memory.
\newblock {\em Neural Computation}, 9(8):1735--1780.

\bibitem[\protect\citename{Imbir}2017]{Imbir17}
Kamil~K Imbir.
\newblock 2017.
\newblock The affective norms for polish short texts ({ANPST}) database
  properties and impact of participants' population and sex on affective
  ratings.
\newblock {\em Frontiers in Psychology}, 8:855.

\bibitem[\protect\citename{Joshi \bgroup et al.\egroup }2020]{Joshi19spanbert}
Mandar Joshi, Danqi Chen, Yinhan Liu, Daniel~S. Weld, Luke Zettlemoyer, and
  Omer Levy.
\newblock 2020.
\newblock {S}pan{BERT}: Improving pre-training by representing and predicting
  spans.
\newblock {\em Transactions of the Association for Computational Linguistics},
  8:64--77.

\bibitem[\protect\citename{Kalchbrenner \bgroup et al.\egroup
  }2014]{Kalchbrenner14}
Nal Kalchbrenner, Edward Grefenstette, and Phil Blunsom.
\newblock 2014.
\newblock A convolutional neural network for modelling sentences.
\newblock In {\em Proceedings of the 52nd Annual Meeting of the Association for
  Computational Linguistics (Volume 1: Long Papers)}, pages 655--665.

\bibitem[\protect\citename{Khanpour \bgroup et al.\egroup }2017]{Khanpour17}
Hamed Khanpour, Cornelia Caragea, and Prakhar Biyani.
\newblock 2017.
\newblock Identifying empathetic messages in online health communities.
\newblock In {\em {Proceedings} of the 8th {International} {Joint} {Conference}
  on {Natural} {Language} {Processing} ({Volume} 2: {Short} {Papers})}, pages
  246--251.

\bibitem[\protect\citename{Kingma and Ba}2015]{Kingma15}
Diederik~P. Kingma and Jimmy Ba.
\newblock 2015.
\newblock Adam: {A} method for stochastic optimization.
\newblock In {\em {Proceedings} of the 3rd {International} {Conference} on
  {Learning} {Representations}}.

\bibitem[\protect\citename{Li \bgroup et al.\egroup }2017]{Li17}
Minglei Li, Qin Lu, Yunfei Long, and Lin Gui.
\newblock 2017.
\newblock Inferring affective meanings of words from word embedding.
\newblock {\em IEEE Transactions on Affective Computing}, 8(4):443--456.

\bibitem[\protect\citename{Liu \bgroup et al.\egroup }2017]{Liu17personality}
Fei Liu, Julien Perez, and Scott Nowson.
\newblock 2017.
\newblock A language-independent and compositional model for personality trait
  recognition from short texts.
\newblock In {\em Proceedings of the 15th {Conference} of the {European}
  {Chapter} of the {Association} for {Computational} {Linguistics}: {Volume} 1,
  {Long} {Papers}}, pages 754--764.

\bibitem[\protect\citename{Mikolov \bgroup et al.\egroup }2013]{Mikolov13nips}
Tomas Mikolov, Ilya Sutskever, Kai Chen, Greg~S Corrado, and Jeff Dean.
\newblock 2013.
\newblock Distributed representations of words and phrases and their
  compositionality.
\newblock In C.~J.~C. Burges, L.~Bottou, M.~Welling, Z.~Ghahramani, and K.~Q.
  Weinberger, editors, {\em Advances in Neural Information Processing Systems
  26}, pages 3111--3119. Curran Associates, Inc.

\bibitem[\protect\citename{Mohammad and Bravo-Marquez}2017a]{Mohammad17starsem}
Saif Mohammad and Felipe Bravo-Marquez.
\newblock 2017a.
\newblock Emotion intensities in tweets.
\newblock In {\em Proceedings of the 6th {Joint} {Conference} on {Lexical} and
  {Computational} {Semantics}}, pages 65--77.

\bibitem[\protect\citename{Mohammad and Bravo-Marquez}2017b]{Mohammad17wassa}
Saif Mohammad and Felipe Bravo-Marquez.
\newblock 2017b.
\newblock {WASSA-2017} shared task on emotion intensity.
\newblock In {\em Proceedings of the 8th Workshop on Computational Approaches
  to Subjectivity, Sentiment and Social Media Analysis}, pages 34--49.

\bibitem[\protect\citename{Mohammad and Kiritchenko}2015]{Mohammad15}
Saif~M Mohammad and Svetlana Kiritchenko.
\newblock 2015.
\newblock Using hashtags to capture fine emotion categories from tweets.
\newblock {\em Computational Intelligence}, 31(2):301--326.

\bibitem[\protect\citename{Mohammad \bgroup et al.\egroup
  }2018]{Mohammad18semeval}
Saif Mohammad, Felipe Bravo-Marquez, Mohammad Salameh, and Svetlana
  Kiritchenko.
\newblock 2018.
\newblock {SemEval-2018 Task 1}: Affect in tweets.
\newblock In {\em Proceedings of the 12th International Workshop on Semantic
  Evaluation}, pages 1--17.

\bibitem[\protect\citename{Pedregosa \bgroup et al.\egroup }2011]{Pedegrosa11}
Fabian Pedregosa, Ga{\"{e}}l Varoquaux, Alexandre Gramfort, Vincent Michel,
  Bertrand Thirion, Olivier Grisel, Mathieu Blondel, Peter Prettenhofer, Ron
  Weiss, Vincent Dubourg, and {others}.
\newblock 2011.
\newblock Scikit-learn: {Machine} learning in {Python}.
\newblock {\em Journal of Machine Learning Research}, 12:2825--2830.

\bibitem[\protect\citename{Pennington \bgroup et al.\egroup
  }2014]{Pennington14}
Jeffrey Pennington, Richard Socher, and Christopher~D. Manning.
\newblock 2014.
\newblock {GloVe}: Global vectors for word representation.
\newblock In {\em Proceedings of the 2014 Conference on Empirical Methods in
  Natural Language Processing}, pages 1532--1543.

\bibitem[\protect\citename{Peters \bgroup et al.\egroup }2018]{Peters2018naacl}
Matthew Peters, Mark Neumann, Mohit Iyyer, Matt Gardner, Christopher Clark,
  Kenton Lee, and Luke Zettlemoyer.
\newblock 2018.
\newblock Deep contextualized word representations.
\newblock In {\em Proceedings of the 2018 Conference of the North American
  Chapter of the Association for Computational Linguistics: Human Language
  Technologies, Volume 1 (Long Papers)}, pages 2227--2237.

\bibitem[\protect\citename{Pinheiro \bgroup et al.\egroup }2017]{Pinheiro17}
Ana~P. Pinheiro, Marcelo Dias, Jo\~{a}o Pedrosa, and Ana~P. Soares.
\newblock 2017.
\newblock Minho {Affective} {Sentences} ({MAS}): {Probing} the roles of sex,
  mood, and empathy in affective ratings of verbal stimuli.
\newblock {\em Behavior Research Methods}, 49(2):698--716.

\bibitem[\protect\citename{Reimers and Gurevych}2018]{Reimers18arxiv}
Nils Reimers and Iryna Gurevych.
\newblock 2018.
\newblock Why comparing single performance scores does not allow to draw
  conclusions about machine learning approaches.
\newblock {\em arXiv:1803.09578 [cs.LG]}.

\bibitem[\protect\citename{Rosenthal \bgroup et al.\egroup }2017]{Rosenthal17}
Sara Rosenthal, Noura Farra, and Preslav Nakov.
\newblock 2017.
\newblock {SemEval-2017 Task 4}: {Sentiment} analysis in {Twitter}.
\newblock In {\em Proceedings of the 11th International Workshop on Semantic
  Evaluation}, pages 502--518.

\bibitem[\protect\citename{Rubin}2007]{Rubin07}
Victoria~L. Rubin.
\newblock 2007.
\newblock Stating with certainty or stating with doubt: Intercoder reliability
  results for manual annotation of epistemically modalized statements.
\newblock In {\em Human Language Technologies 2007: The Conference of the North
  American Chapter of the Association for Computational Linguistics; Companion
  Volume, Short Papers}, pages 141--144.

\bibitem[\protect\citename{Srivastava \bgroup et al.\egroup
  }2014]{Srivastava14}
Nitish Srivastava, Geoffrey~E Hinton, Alex Krizhevsky, Ilya Sutskever, and
  Ruslan Salakhutdinov.
\newblock 2014.
\newblock Dropout: {A} simple way to prevent neural networks from overfitting.
\newblock {\em Journal of Machine Learning Research}, 15(1):1929--1958.

\bibitem[\protect\citename{Strapparava and Mihalcea}2007]{Strapparava07}
Carlo Strapparava and Rada Mihalcea.
\newblock 2007.
\newblock {SemEval-2007 Task} 14: Affective text.
\newblock In {\em Proceedings of the 4th International Workshop on Semantic
  Evaluations}, pages 70--74.

\bibitem[\protect\citename{Vaswani \bgroup et al.\egroup }2017]{Vaswani17}
Ashish Vaswani, Noam Shazeer, Niki Parmar, Jakob Uszkoreit, Llion Jones,
  Aidan~N Gomez, {\L}ukasz Kaiser, and Illia Polosukhin.
\newblock 2017.
\newblock Attention is all you need.
\newblock In I.~Guyon, U.~V. Luxburg, S.~Bengio, H.~Wallach, R.~Fergus,
  S.~Vishwanathan, and R.~Garnett, editors, {\em Advances in Neural Information
  Processing Systems 30}, pages 5998--6008. Curran Associates, Inc.

\bibitem[\protect\citename{Yang \bgroup et al.\egroup }2019]{Yang19xlnet}
Zhilin Yang, Zihang Dai, Yiming Yang, Jaime Carbonell, Russ~R Salakhutdinov,
  and Quoc~V Le.
\newblock 2019.
\newblock {XLNet}: {Generalized} autoregressive pretraining for language
  understanding.
\newblock In H.~Wallach, H.~Larochelle, A.~Beygelzimer, F.~d'~Alch\'{e}-Buc,
  E.~Fox, and R.~Garnett, editors, {\em Advances in Neural Information
  Processing Systems 32}, pages 5753--5763. Curran Associates, Inc.

\end{thebibliography}
